%% file: main.tex
\documentclass[11pt,a4paper]{article}
\usepackage{times,latexsym}
\usepackage{url}
\usepackage[T1]{fontenc}

\usepackage{tacl2021v1}
\usepackage{inconsolata}

\usepackage{tcolorbox}
\usepackage{changepage}
\usepackage{booktabs}
\usepackage{graphicx}
\usepackage{subcaption}
\usepackage{tabularx}
\usepackage{multirow, multicol}
\newcolumntype{P}[1]{>{\centering\arraybackslash}p{#1}}

\usepackage{todonotes}
\usepackage{tcolorbox}
\usepackage{changepage}
\usepackage{booktabs}
\usepackage{graphicx}
\usepackage{subcaption}
\usepackage{tabularx}
\usepackage{multirow, multicol}
\usepackage{xcolor}  
\usepackage{colortbl}
\usepackage{tikz}
\definecolor{lightgray}{gray}{0.9}
\definecolor{lightred}{rgb}{0.98, 0.81, 0.82}
\newcolumntype{P}[1]{>{\centering\arraybackslash}p{#1}}
\newenvironment{labelledp}[1]
  {\par\refstepcounter{equation}\label{#1}}
  {\par}

\usepackage{xspace,mfirstuc,tabulary}

\taclpubformattrue
\newif\iftaclinstructions
\taclinstructionsfalse 
\iftaclinstructions
\renewcommand{\confidential}{}
\renewcommand{\anonsubtext}{(No author info supplied here, for consistency with
TACL-submission anonymization requirements)}
\newcommand{\instr}
\fi

\iftaclpubformat 

\else

\fi

\title{Multimodal Large Language Models to Support Real-World Fact-Checking}

\author{
${\textbf{Jiahui Geng}}$,
${\textbf{Yova Kementchedjhieva}}$, 
${\textbf{Preslav Nakov}}$,
${\textbf{Iryna Gurevych}}$  \\
 Mohamed bin Zayed University of Artificial Intelligence \\
\{jiahui.geng, yova.kementchedjhieva,preslav.nakov,iryna.gurevych\}@mbzuai.ac.ae 
}

\date{}

\begin{document}
\maketitle

\input{sections/00_abstract}
\input{sections/01_introduction}
\input{sections/02_background}
\input{sections/03_methodology}
\input{sections/04_evaluation}

\input{sections/06_limitations}
\input{sections/05_conclustion}
\bibliography{tacl2021,acl}
\bibliographystyle{acl_natbib}

\appendix

\input{sections/07_appendix}

\end{document}

%% file: sections/00_abstract.tex
\begin{abstract}
Misinformation poses a real-world threat, especially when combined with misleading images. Multimodal large language models (MLLMs), which combine image comprehension with the rich knowledge and explanatory capability of language models, have become tools for humans to process vast information. However, their capacity and limitations as multimodal tools in assisting with fact-checking remain understudied. Here is aim to bridge this gap. In particular, we propose a framework for systematically assessing the capacity of current multimodal models to facilitate real-world fact-checking. Our methodology is evidence-free, leveraging only these models' intrinsic knowledge and reasoning capabilities. By designing prompts that extract models' predictions, explanations, and confidence levels, we conduct a detailed analysis of the model's accuracy, bias, and other key factors. We empirically find that (1) GPT-4V exhibits surprising performance across various datasets, with an accuracy rate exceeding $80\%$, and is capable of providing impressive explanations, and (2) even with the aid of prompt ensembles and in-context learning, open-source models significantly lag in performance. However, they show potential in remembering checked claims and reasoning out manipulated images. We also summarize the failure reasons, which contributes to strategies for future improvements. Our study offers insights into leveraging MLLMs to combat multimodal misinformation.


\end{abstract}

%% file: sections/01_introduction.tex
\section{Introduction} \label{sec:intro}
Misinformation is a significant challenge on the internet, especially with regard to multimodal claims, which combine text, images, videos, and other media types~\cite{mubashara2023multimodal}. The visual component in these cases could be manipulated or used out-of-context (OOC) to make a false claim~\cite{Huh_2018_ECCV,luo-etal-2021-newsclippings,aneja2023cosmos,mocheg}. In such cases, fact-checkers and the tools they employ need to be able to handle multiple modalities.

\begin{figure}[t]
    \centering
    \includegraphics[width=\linewidth]{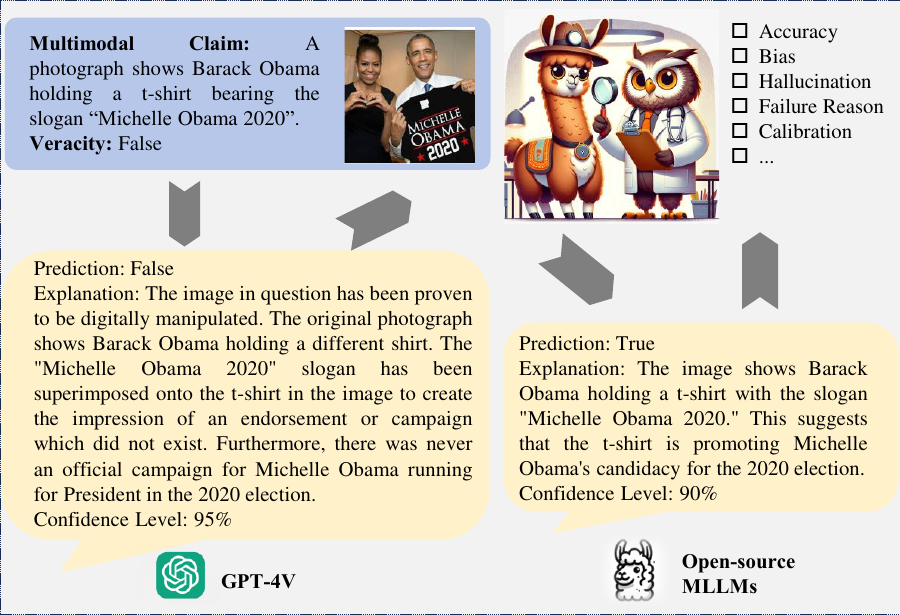}
    \caption{Illustration of our proposed framework to evaluate the capability of MLLMs as fact-checkers. Initially, we collect their responses to multimodal claims, encompassing predictions, explanations, and confidence levels. We then assess their performance across dimensions, including accuracy, bias, and their failure reasons.}
    \label{fig:demo}
\end{figure}

Large language models (LLMs) store extensive information, which exceeds what any individual human can know and is accessible in a far more human-friendly way than any search engine can offer~\cite{zhao2023survey,hu2023large}. As such, they can be a powerful tool in the hands of fact-checkers, who often seek extra factual knowledge to verify a claim~\cite{nakov2021automated}. 
The application of LLMs to fact-checking has been extensively studied in a text-only setting, wherein a model receives as input both a textual claim, with or without a set of evidence documents, and needs to reason to determine the veracity of the claim~\cite{chen2022gere,pan-etal-2023-fact,li2023self}. LLMs have increasingly strong understanding and reasoning capabilities, so using them to sift through evidence documents when verifying a claim is proving very effective \cite{chen2022gere,li2023self}. Yet, this pipeline approach requires an extra step of evidence retrieval, which is computationally heavy and error-prone, while it overlooks the vast amount of knowledge stored in the parameters of LLMs~\cite{hu2023read}. We hypothesize that multimodal large language models (MLLMs) trained on large amounts of data can serve as a sufficient substitute for the retrieval of evidence and perform fact-checking in an evidence-free fashion, i.e., relying solely on their parametric knowledge.
 
Despite their great promise, MLLMs have limitations as well, specifically with respect to factuality \cite{fu2023mme,liu2023mmbench}. Models may produce hallucinations \cite{huang2023survey,li-etal-2023-evaluating} and suffer from a lack of knowledge or exhibit biases~\cite{hu2023large,wang2023survey}. 
To understand MLLMs’ potential to support fact-checking real-world claims, we propose the evaluation framework illustrated in Figure~\ref{fig:demo}. It involves querying models for predictions, explanations, and confidence scores on multimodal check-worthy claims. Based on the data obtained, we can further assess the models' capabilities from various perspectives, addressing the following research questions:


\begin{itemize}
    \item \textbf{RQ1 Performance Evaluation:}  How good are MLLMs in identifying multimodal misinformation? Can they perform accurate reasoning? 
    \item \textbf{RQ2 Improving Approaches:} How can we effectively boost the model's fact-checking capabilities? Can the improvements be reflected in the reasoning, beyond accuracy metrics?
    \item \textbf{RQ3 Taxonomy of Failure Reasons:}  What are the typical errors for MLLMs employed as multimodal fact-checkers?
\end{itemize}

Our investigation spans an array of vision-language models, including GPT-4V, MiniGPT-v2~\cite{chen2023minigptv2} and LLaVA-1.5~\cite{liu2023visual}, which stand as leaders in this domain. We evaluate these models across three existing multimodal fact-checking datasets: Fauxtography~\cite{zlatkova-etal-2019-fact}, COSMOS~\cite{aneja2023cosmos}, MOCHEG~\cite{mocheg}, and one new dataset, which build from recent check-worthy claims to mitigate the risk of data contamination. We use prompt ensembles (PE) and in-context learning (ICL) to help improve the open-source MLLMs. We do not use fine-tuning because models are required to explain its reasoning process and provide confidence scores, while high-quality annotated data is scarce. Our study finds that MLLMs display a range of fact-checking capabilities. GPT-4V, in particular, excels in our tests, exhibiting high accuracy, useful explanations, and good calibration. The accuracy on various datasets generally reaches around $80\%$. MiniGPT-v2 lacks explanatory and uncertainty-reporting capabilities. LLaVAs can follow instructions, and both PE and ICL enhance their performance to varying degrees, with the latter providing greater improvement. On Fauxtography and COSMOS, ICL increases the Overall Accuracy from 52.3\% to 62.3\% and from 67.7\% to 76.5\%, respectively. The majority baseline for both datasets is 50\%.

In summary, we systematically evaluate existing multimodal models to determine their capability to support real-world fact-checking. Our approach is more comprehensive than prior studies, which are limited to specific aspects such as face spoofing, forgery detection, or out-of-context detection~\cite{shi2024shield,qi2024sniffer}. We formulate several research questions to evaluate these models based on accuracy, bias, reasoning capabilities, and errors across a variety of datasets. Our work fills a research gap, providing a comprehensive overview of the domain and shedding light on leveraging MLLMs to combat multimodal misinformation.

%% file: sections/02_background.tex
\section{Related Work}\label{sec:related}
\paragraph{LLMs for Text-Only Fact-Checking}
\citet{caramancion2023news} compared the performance of models such as ChatGPT 3.5 and ChatGPT 4.0 in news fact-checking.FactLLaMA~\cite{cheung2023factllama} integrates external evidence into the instruct-tuning process to enhance the model's ability to leverage evidence for predictions. Studies have also focused on leveraging the capabilities of LLMs to develop automated frameworks for decomposing claims, generating queries, and predicting based on gathered evidence, thereby augmenting the interpretability of the fact-checking process ~\cite{pan-etal-2023-fact,li2023self}. However, these works often focused solely on prediction accuracy. Our work analyzes the model's fact-checking capabilities from three perspectives: predictions, explanations, and confidence.

\paragraph{Multimodal Fact-Checking}
Multimodal misinformation primarily falls into two categories: one involves textual claims about manipulated content across different modalities, while the other pertains to out-of-context misinformation, featuring unaltered images, audio, or videos in misleading contexts~\cite{luo-etal-2021-newsclippings,aneja2023cosmos,mocheg}. \citet{shi2024shield} proposed a benchmark to evaluate the performance of MLLMs in detecting face spoofing and forgery. \citet{qi2024sniffer} introduced Sniffer, an MLLM designed for OOC detection and explanation, utilizing fine-tuning to boost the accuracy in identifying celebrities and clarifying inconsistencies between text and images. In our work, we posit that evidence is parametrically encoded within the models, enabling us to bypass the evidence retrieval phase and directly engage in predictions and explanations for the real-world claims.

%% file: sections/03_methodology.tex
\section{Evaluation Framework}\label{sec:method}
We propose an evaluation framework that includes datasets, prompts, and evaluation metrics to address the research questions.

\subsection{Datasets}
Due to constraints posed by the GPT-4V API, such as decreased query speeds after exceeding a daily limit, we sample data from various datasets and preprocess them to fit our evaluation framework. Additionally, we develop a new dataset to ensure the model has not encountered these data.

\paragraph{Fauxtography} is a multimodal fact-checking dataset sourced from the websites Snopes and Reuters~\cite{zlatkova-etal-2019-fact}.  We select a random subset of 400 entries, ensuring a balanced composition of 200 true and 200 false instances. 

\paragraph{COSMOS} is a dataset built from Snopes and News Outlets, and tailored for OOC detection~\cite{aneja2023cosmos}. We randomly select 240 true and 240 false samples from the test split. Each image in this dataset comes with two captions, and we manually select one caption per image to ensure it aligns with our criteria.


\paragraph{MOCHEG} is originally based on textual claims from Snopes and PolitiFact, with associated images serving as evidence~\cite{mocheg}. This implies that a single claim can correspond to multiple images, many of which are not explicitly required for the claim to be verified. We first identify multimodal claims through keywords such as \textit{photograph}, \textit{image}, etc. Subsequently, we manually select images that were explicitly requested for verification and filter out those with overlaying words like \textit{fake}, \textit{misleading}, or \textit{miscaptioned}. This procedure results in a total of 504 data entries.



\paragraph{Post-4V} We collect new data from Snopes, focusing on articles published after the release of GPT-4V (September 26, 2023). We apply the same filtering methodology used in MOCHEG, resulting in a dataset of 186 samples.
\label{sec:annotation}

\begin{table}[t]
\centering
\scriptsize
\begin{tabular}{p{0.5cm}<{\centering}p{1.2cm}<{\centering}p{1.2cm}<{\centering}p{1.2cm}<{\centering}p{1.2cm}<{\centering}}
\toprule
 & Fauxtography & COSMOS & MOCHEG & Post-4V \\
\midrule
True  &  200  &    240 &     267 &      81 \\
False &   200&      240&     237 &      105\\
\midrule
Total &   400&      480&     504&      186\\
\bottomrule
\end{tabular}
\caption{Statistics of datasets in our evaluation.}
\label{tab:stats}
\end{table}

\subsection{Evaluation Prompt}
\label{sec:eval_prompt}
We simultaneously obtain the predictions, explanations, and confidence levels from MLLMs with the prompt below. This prompt is formulated following the example of prompts found in related fact-checking work ~\cite{pan-etal-2023-fact,min-etal-2023-factscore}, while for acquiring explanations and confidence estimates we refer to \citet{xiong2023can}. We collect their verbalized confidence as it is increasingly used for decision-making during human-machine collaboration~\cite{geng2023survey}.
\begin{adjustwidth}{0.5cm}{0.5cm}
\footnotesize
\texttt{\\Is it true that "\textbf{CLAIM}"? True or False? Use the following format to provide your answer: \\}
\texttt{ Prediction:  [True or False] \\
Explanation: [put your evidence and step-by-step reasoning here] \\
Confidence Level: [please show the percentage] \vspace{\baselineskip} \\}
\noindent \texttt{Note: The confidence level indicates the degree of certainty you have about your answer and is represented as a percentage. For instance, if your confidence level is $80\%$, it means you are $80\%$ certain that your answer is correct and there is a $20\%$ chance that it may be incorrect.}
\end{adjustwidth}

\input{results/04_comprehensive_table}

\subsection{Evaluation Metrics}

\paragraph{Response Types}

Based on the three components (prediction, explanation, and confidence level) in the model's response, we categorize them into four types.
\textit{Others} refer to instances where the model fails to respond according to our requirements, missing any component. \textit{Uncertain} denotes cases where the model explicitly expresses uncertainty, with statements such as ``cannot confirm'', ``cannot verify'', or ``cannot determine'', etc. The remaining samples are grouped as \textit{True} or \textit{False} according to the model's prediction. Overall, the numbers of these four indicators reveal the model's ability to follow instructions, express uncertainty, and inherent bias in predicting True and False.


\paragraph{Accuracy Metrics}
We design two accuracy metrics to reflect the model's performance. One is~\textit{True\&False Accuracy}, measuring the accuracy of samples that are solely classified as True or False. This is important because users usually do not rely on the model's response when the model clearly expresses uncertainty or does not follow instructions. The other metric, named \textit{Overall Accuracy}, calculates the proportion of responses that contain correct predictions across all samples. This facilitates comparison between different approaches.

\section{Experimental Setups}
\label{sec:setup}
\begin{figure}[!htbp]
    \centering
    \includegraphics[width=0.8\linewidth]{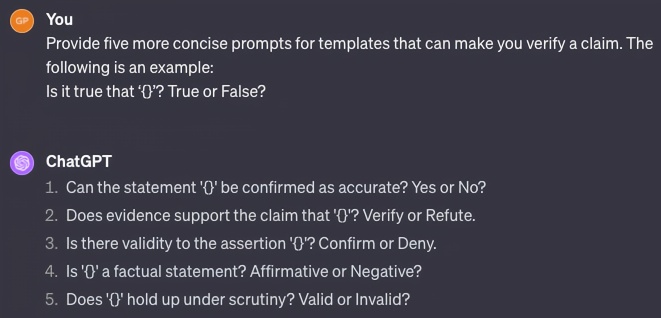}
    \caption{Prompts obtained from ChatGPT that are used in prompt ensembles experiments.}
    \label{fig:prompts}
\end{figure}

We use OpenAI's API to collect responses from GPT-4V (gpt-4-1106-preview). In addition, we experiment with open-source MLLMs, including LLaVA-1.5 (7b and 13b, ~\citealt{liu2023visual,liu2023improved}), and MiniGPT (v2,~\citealt{chen2023minigptv2}). All parameters are set to the default values, with \textit{max\_tokens=300} for GPT-4V. These models are notably representative and competitive in a multimodal setup~\cite{fu2023mme,liu2023mmbench}. We further explore two approaches that do not require fine-tuning to enhancing the model's performance:

\paragraph{Prompt Ensembles (PE)} involve using a variety of prompts for the same task and aggregating model's responses to produce more accurate and less biased predictions. We employ ChatGPT to generate 5 more semantically similar yet distinct prompts, as depicted in Figure~\ref{fig:prompts}. The part about explanations and confidence scores is identical to the original prompt. With these prompts we collect six responses for each claim. Then, we conduct a vote to determine the final response type. If the two highest-scoring response types receive the same number of votes, we consider the model uncertain.

%


\paragraph{In-Context Learning (ICL)} enables a model to learn tasks by directly observing examples within the prompt, bypassing the need for prior explicit training. Debunking articles, which are lengthy and often contain irrelevant descriptions, are not used here. Instead, we utilize responses from GPT-4V to claims not included in our test datasets as instructive examples. To overcome LLaVA's limitation in handling multiple image inputs, we vertically stack different images with a 20-pixel separation. The relationship between each image and its corresponding claim is explicitly described, such as "For the first picture, claim: $\cdots$", as shown in the prompt below. We perform 1 and 2-shot learning, showing the model one and two examples, respectively, referred to as ICL-1 and ICL-2. We use four pairs of examples as demonstrations. For ICL-1, all examples are False, while for ICL-2,  the first claim is always False, and the second is True. They are detailed illustrated in Figure~\ref{fig:icl} in Appendix.

\begin{labelledp}{prompt_1}
\begin{adjustwidth}{0.5cm}{0.5cm}
\footnotesize
\texttt{\\Use the following format to answer whether the claim regarding the pictures is True or False: \\
\textbf{SAME FORMAT AS IN THE ORIGINAL PROMPT.}\\}

\noindent \texttt{For the first picture,  claim: "\textbf{CLAIM1}" \\
Prediction: False \\
Explanation: XXX  \\
Confidence Level: XXX\\}

\noindent \texttt{For the second picture,  claim: "\textbf{CLAIM2}" \\
Prediction: True \\
Explanation: XXX \\
Confidence Level: XXX \\}

\noindent \texttt{For the third picture, claim: "\textbf{CLAIM}"}
\end{adjustwidth}
\end{labelledp}

%% file: results/04_comprehensive_table.tex
\begin{table*}[t]
\centering
\scriptsize
\setlength{\tabcolsep}{3pt}
\begin{tabular}{p{1.7cm}p{0.35cm}<{\centering}p{0.35cm}<{\centering}p{0.2cm}<{\centering}p{0.3cm}<{\centering}>{\columncolor{lightgray}}p{0.45cm}<{\centering}>{\columncolor{lightgray}}p{0.5cm}<{\centering}|p{0.35cm}<{\centering}p{0.35cm}<{\centering}p{0.2cm}<{\centering}p{0.3cm}<{\centering}>{\columncolor{lightgray}}p{0.45cm}<{\centering}>{\columncolor{lightgray}}p{0.5cm}<{\centering}|p{0.35cm}<{\centering}p{0.35cm}<{\centering}p{0.2cm}<{\centering}p{0.3cm}<{\centering}>{\columncolor{lightgray}}p{0.45cm}<{\centering}>{\columncolor{lightgray}}p{0.5cm}<{\centering}|p{0.35cm}<{\centering}p{0.35cm}<{\centering}p{0.2cm}<{\centering}p{0.3cm}<{\centering}>{\columncolor{lightgray}}p{0.45cm}<{\centering}>{\columncolor{lightgray}}p{0.5cm}<{\centering}}
\toprule
     & \multicolumn{6}{c}{\textbf{Fauxtography}} & \multicolumn{6}{c}{\textbf{COSMOS}} & \multicolumn{6}{c}{\textbf{MOCHEG}} & \multicolumn{6}{c}{\textbf{Post-4V}} \\ \midrule 
  & T    & F    & U    & O   & \cellcolor{white}{T\&F} & \cellcolor{white}{All}   & T   & F   & U  & O  & \cellcolor{white}{T\&F} & \cellcolor{white}{All}  & T   & F   & U  & O  & \cellcolor{white}{T\&F} & \cellcolor{white}{All}   & T   & F  & U  & O  & \cellcolor{white}{T\&F} & \cellcolor{white}{All} \\
\midrule
 GPT-4V     &  158    &   195   &   29  & 18 &  \textbf{81.9} & \textbf{82.1}   & 179  &  204   & 83  & 14 &  \textbf{86.2} & \textbf{80.0}  &  216   & 223   & 37  & 28   &  \textbf{87.2} & \textbf{83.5}  & 54     &  98   &  26  &  8  &      \textbf{79.6} & \textbf{73.7}  \\
MiniGPT       &  -    & -     & -     &  -  &   {-} & {55.5}   &   -  &   -  &   - &  -  &   - & 62.1   &   -  &  -   & -   & -   &  - & 54.3  &  -   &   - &  -  &  -  &  - & 55.9   \\
LLaVA(7b)     &   337   &  6    &   1  &   56  &   {53.4} & {46.3}    & 449    &  7   &  4  &  20  &  52.0 & 50.0    &  409  &   6  &  0  &  89  &   54.2 & 44.8   &  157   &  1  &  3  &  25  &  42.0 & 37.1   \\
LLaVA(13b)  & 286    & 98     & 3     &  13   &  {54.4} & {52.3}     &   331  &  134   & 13   &  2  &  69.7 & 67.7    &  363   &  112   &  28  &  1  &   61.5 & 58.1   &  116   & 65   &  2  &  3  &   56.4 & 55.9  \\ 
LLaVA+PE & 244 & 153 & 2 & 1 & 57.1 & 54.7 &  275 & 204 & 0 & 1 & 76.3 & 71.7
& 290 & 214 & 0 & 0 & 59.9 & 58.1 & 85 & 101 &  0 & 0 & 56.9 & 56.1 \\
 LLaVA+ICL-1 & 228 & 159 & 6 & 8 & \textbf{61.8} & \textbf{62.3} & 293 & 175 & 9 & 3 & 74.8 & 74.1 &
 254 & 240 & 7 & 4 & \textbf{62.4} & \textbf{62.5} & 79 & 101 & 5 & 1 & 58.7 & 57.5 \\
LLaVA+ICL-2 & 186 & 188 & 8 & 18 & 61.6 & 61.7 & 247 & 215 & 12 & 7 & \textbf{77.3} & \textbf{76.5} &
195 & 286 & 10 & 13 & 60.2 & 60.4 & 48 & 122 & 8 & 8 & \textbf{62.1} & \textbf{61.4} \\
Majority    &  -   &  -    &  -    &   - &  {-}  &  {50.0}   &  -   & -   &   - &  &     -   &  50.0  &  -   & -   &  -  &  -   &   -  &  53.0  &   - & -  &  -   &   -&   -  & 56.5 \\
\bottomrule
\end{tabular}
\caption{Performance of various models and approaches. \textit{T: True, F: False, U: Uncertain, O: Others, T\&F: True\&False Accuracy, All: Overall Accuracy, PE: Prompt Ensembles, ICL: In-Context Learning.}  The majority-class accuracy is established in the last row. }
\label{Performance of various models.}
\label{tab:comprehensive}
\end{table*}

%% file: sections/04_evaluation.tex
\section{Experimental Results}\label{sec:evaluation}
We comprehensively analyze the data we collected in Section~\ref{sec:comprehensive}. In Section~\ref{sec:explain}, we provide a detailed analysis of the model explanations, focusing on their step-by-step reasoning, and discuss our taxonomy of failure reasons. Section~\ref{sec:ablation} is dedicated to ablation studies.

\subsection{Comprehensive Results}
\label{sec:comprehensive}
The comprehensive results are displayed in Table~\ref{tab:comprehensive}. The first four columns for each dataset represent the number of different types of responses, while the following are True\&False and Overall Accuracy, respectively. We have highlighted the highest metrics and the highest ones among the open-source models. Hereafter, we omit the version numbers of the open-source models and refer to them simply as LLaVA(7b), LLaVA(13b), and MiniGPT. Both PE and ICL experiments are based on LLaVA(13b). LLaVA+PE involves voting based on 6 responses; if the counts of True and False are equal, then it is an uncertain case. LLaVA+ICL-1 and LLaVA+ICL-2 represent the average results across four sets of demonstrations, we have rounded the number of different responses. Detailed outcomes for PE and ICL are presented in Table~\ref{tab:pe} and~\ref{tab:icl} in Appendix.

Overall, GPT-4V demonstrates surprising accuracy, with both accuracy metrics exceeding 80\% on the three public datasets. The accuracy decrease is not particularly significant on Post-4V, with the True\&False Accuracy equals $79.6\%$. In addition, these values are mostly higher than Overall Accuracy, suggesting that GPT-4V exhibits higher precision when it responds with confidence.  There are more False cases than True ones. The high number of \textit{Others} is primarily due to GPT-4V safety alignment. This results in frequent replies such as "I'm sorry, but I cannot assist with this request." or "This content may violate our content policy." Additionally, GPT-4V expresses uncertainty the most.

In contrast, MiniGPT cannot provide explanations and confidence, while LLaVA(7b) shows a strong bias, almost always responding True with this prompt. LLaVA(13b) shows improved accuracy across all datasets, better adherence to instructions, and more frequent expressions of uncertainty. We observe that the two approaches, PE and ICL, can enhance the model's performance to varying degrees. As to PE, there is an obvious improvement in the COSMOS dataset, with increases of 6.6 and 4.0 in two accuracy metrics, respectively. However, the enhancement on other datasets is very limited. ICL shows a greater impact. On the COSMOS dataset, LLaVA+ICL-1 brings increases of 7.6 and 9.0 in both metrics. On MOCHEG, the improvement is minimal, with gains of 0.9 and 4.4, respectively.


\begin{figure}[t]
    \centering
    \includegraphics[width=\linewidth]{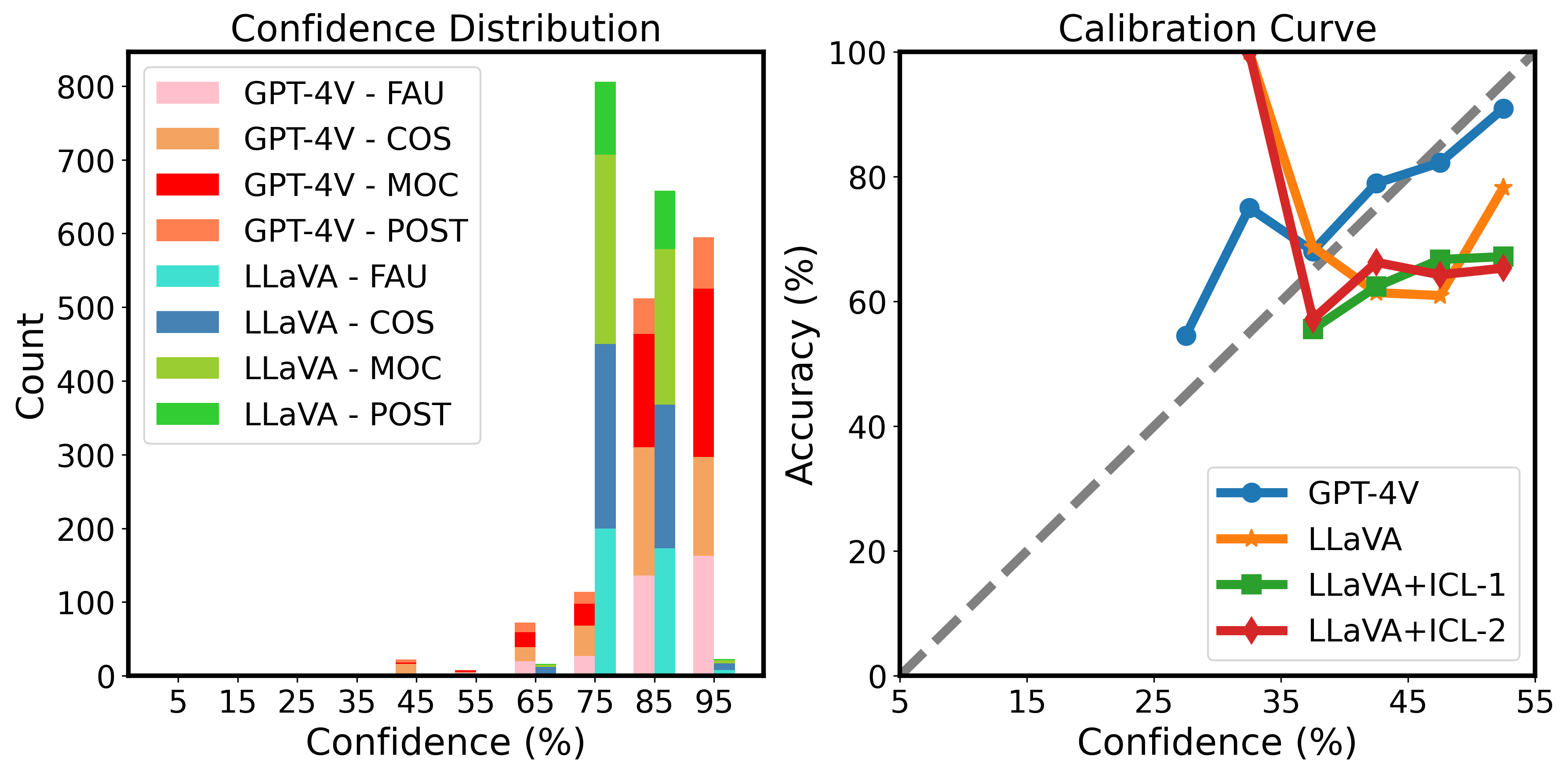}
    \caption{The left graph illustrates the confidence score distribution of GPT-4V and LLaVA(13b), and the right graph presents their calibration curves. \textit{FAU: Fauxtography, COS: COSMOS, MOC: MOCHEG, POST: Post-4V.}}
    \label{fig:confidence}
\end{figure}

\begin{figure*}[!ht]
    \centering
    \includegraphics[width=0.95\textwidth]{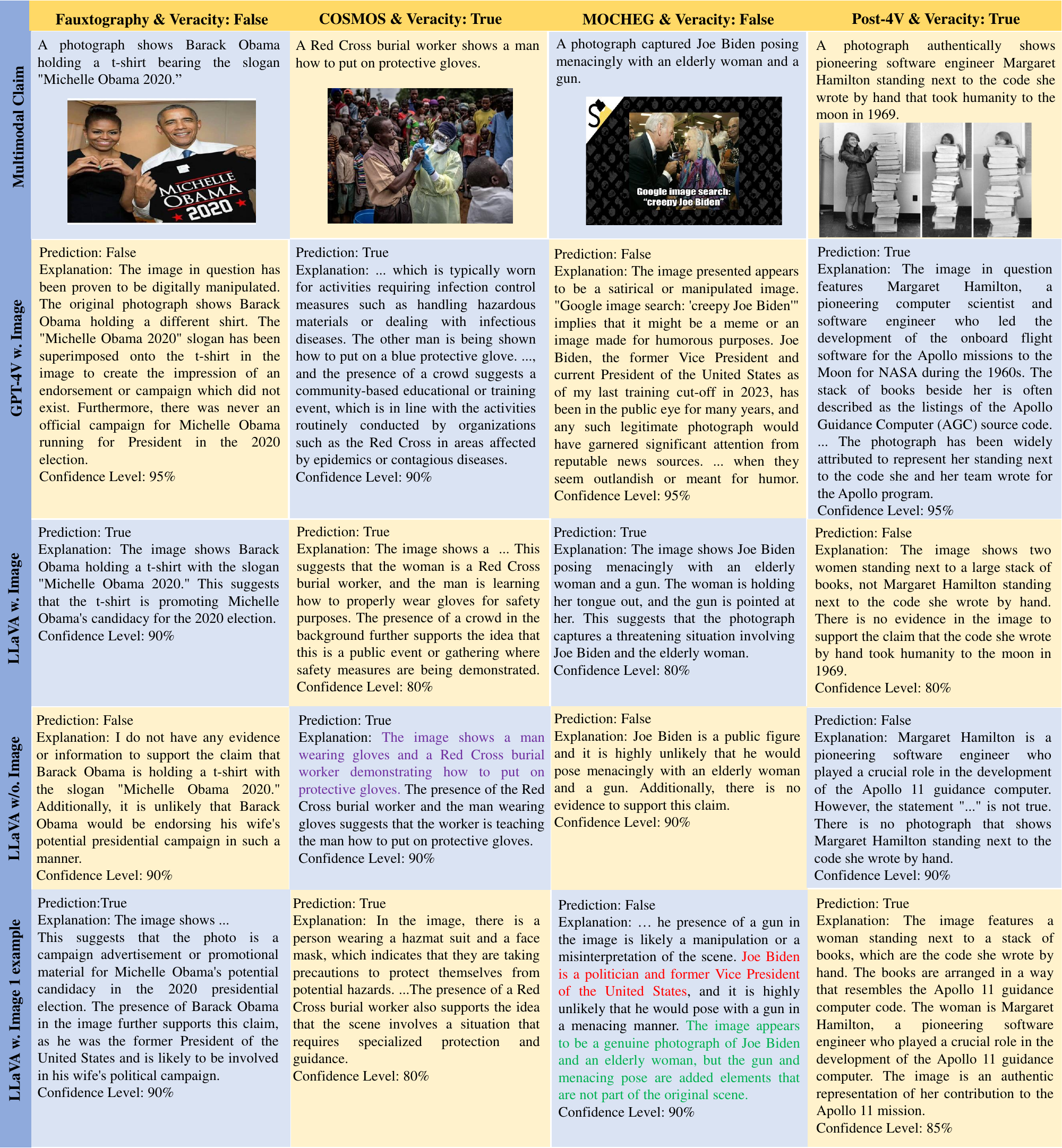}
    \caption{Sampled fact-checking responses from different models and approaches. The first row shows the claim source and its veracity. The second row includes multimodal claims, and the subsequent four rows feature responses from GPT-4V, LLaVA(13b), LLaVA(13b) without image input, and LLaVA+ICL-1 (using the first demonstration), respectively. Purple text indicates hallucinations by the model when no images are present; red text shows outdated knowledge, and green text displays the model's analysis of image manipulation.}
    \label{fig:part-examples}
\end{figure*}

\paragraph{Calibration} The left part of Figure~\ref{fig:confidence} displays the confidence distributions of GPT-4V and LLaVA(13b), breaking down the data into 10\% confidence intervals. For instance, a value of 95 corresponds to confidence levels within the $(90, 100]$ bracket. GPT-4V's confidence scores are largely clustered in the $(80, 100]$ interval, whereas LLaVA(13b)'s scores are more frequently found in the $(70, 90]$ range. The right calibration curves reveal that both GPT-4V and LLaVA exhibit a positive correlation between confidence levels and accuracy, with GPT-4V's calibration closely mirroring the ideal curve (illustrated by the dashed line), indicating well-calibrated confidence scores. Conversely, LLaVA(13b)'s curve suggests a propensity for overconfidence. We find that ICL does not result in better calibration of LLaVA's verbalized confidence as their curves are very close.



\subsection{Explanation-based Analysis}
\label{sec:explain}
\begin{figure}[!t]
    \centering
    \includegraphics[width=\linewidth]{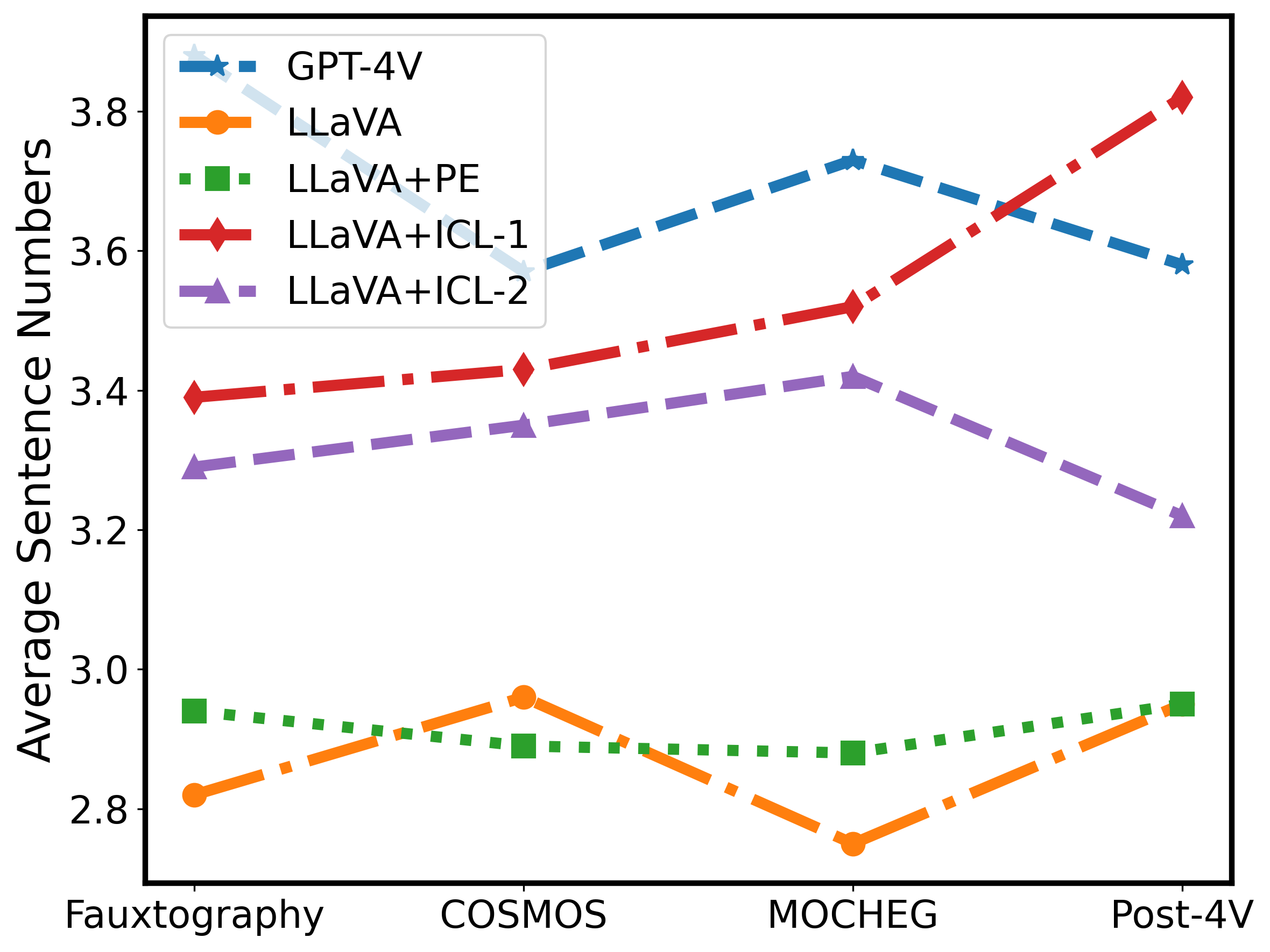}
    \caption{Average number of sentences in explanations across different models and settings. GPT-4V generates the longest explanations except on Post-4V. With one example, ICL-1 significantly increases the average explanation length.}
    \label{fig:sent_nums}
\end{figure}

\begin{figure}[t]
    \centering
    \includegraphics[width=\linewidth]{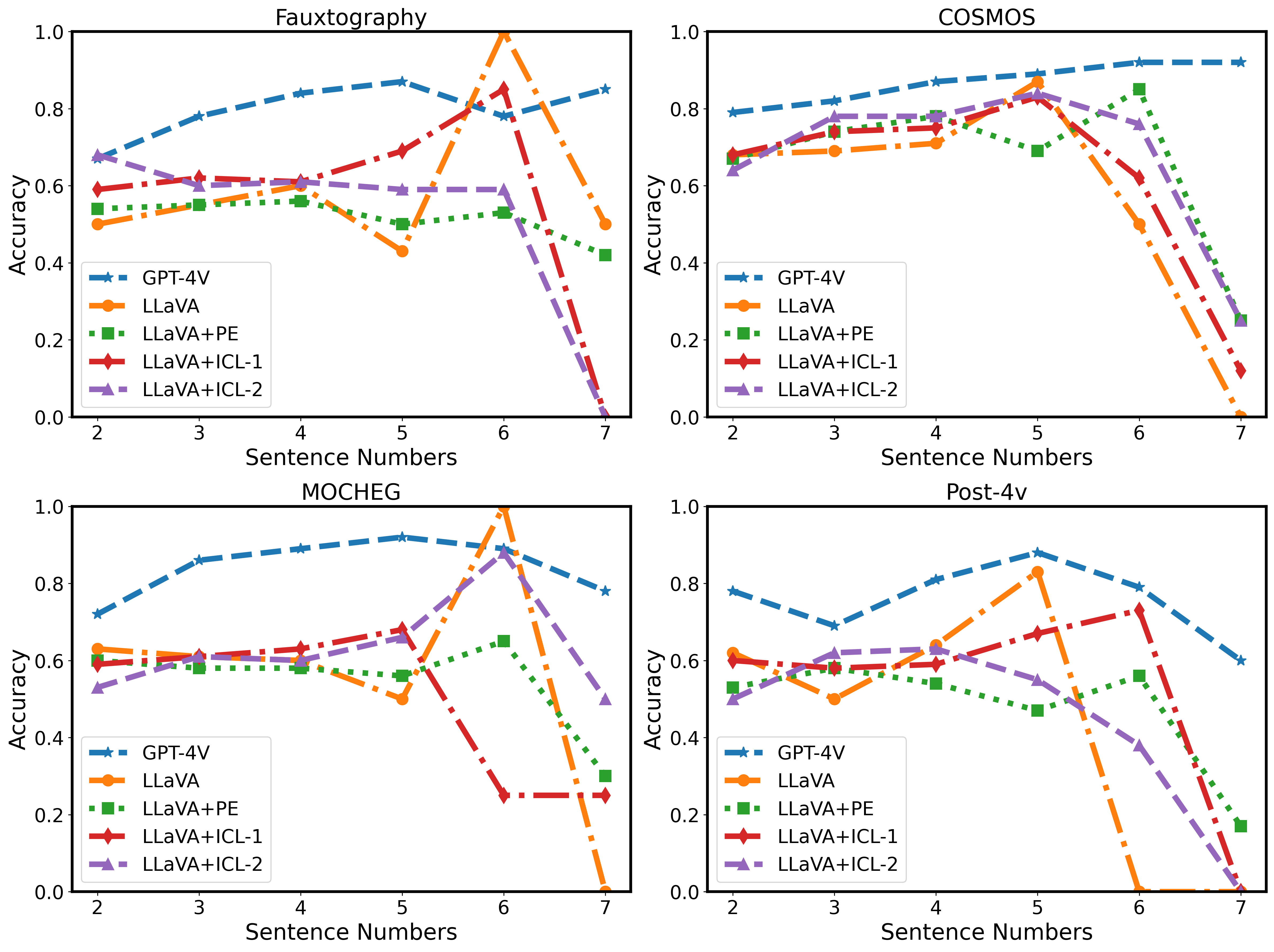}
    \caption{The relationship between response's explanation length and the response accuracy. Generally, accuracy increases with the length of the sentences, peaking at around 5 or 6 sentences, after which it significantly declines.}
    \label{fig:acc_vs_length}
\end{figure}

 \subsubsection{Case Studies}
 Figure~\ref{fig:part-examples} displays totally 4 cases with the original multimodal claims and corresponding responses . Rows labeled "GPT-4V w. Image," "LLaVA w. Image," and "LLaVA w. Image 1 example" correspond to the results of GPT-4V, LLaVA(13b), and LLaVA+ICL-1, respectively. It can be observed that GPT-4V's explanation contains a richer set of facts. For instance, it explicitly states, "there was never an official campaign for Michelle Obama in the 2020 election," in the 1\textsuperscript{st} case, and in the 4\textsuperscript{th} case, it provides detailed background information about Margaret Hamilton, including her role in "leading the development of onboard flight software for the Apollo mission." Moreover, GPT-4V adeptly incorporates details from the images, such as referencing a "Google image search: Creepy Joe Biden" in the 3\textsuperscript{rd} image. 
 
 However, LLaVA(13b) lacks the capability to counteract manipulated images, as evidenced in the 1\textsuperscript{st} and 3\textsuperscript{rd} cases. Additionally, LLaVA(13b)'s explanations are typically brief, often limited to a description of the image followed by a conclusion about its veracity. It shows a lack of effective reasoning, rendering the explanations less comprehensive than GPT-4V. We find ICL enhances LLaVA(13b)'s fact-checking capabilities. For instance, the model accurately analyzes the 3\textsuperscript{rd} image, with "a genuine photograph of Joe Biden and an elderly woman, but the gun and menacing pose are added elements that are not part of the original scene." In the 4\textsuperscript{th} case, LLaVA+ICL-1 provides a more detailed description of Margaret Hamilton’s background and states that "the image is an authentic representation of her contribution." However, it still fails in the 1\textsuperscript{st} case with ICL.

\begin{figure}[t]
    \centering
    \includegraphics[width=\linewidth]{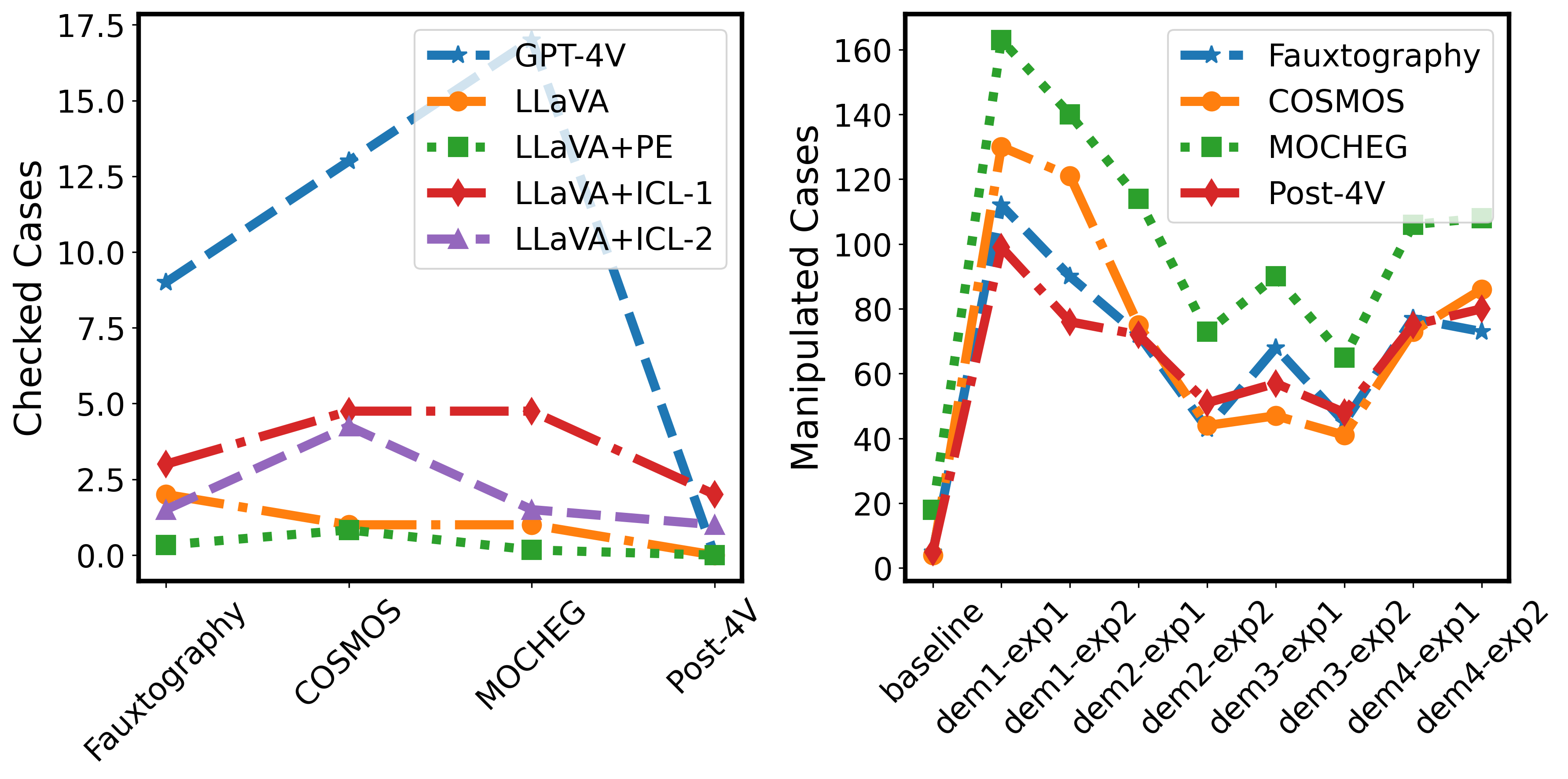}
    \caption{The left graph presents the number of checked cases across various datasets and approaches, with the numbers for LLaVA+PE, LLaVA+ICL-1, and LLaVA+ICL-2 representing average counts. The graph on the right illustrates the comparison of the quantity of manipulated cases identified when using and not using  the examples. The baseline is LLaVA(13b).}
    \label{fig:keywords}
\end{figure}

\begin{figure*}[!ht]
    \centering
    \includegraphics[width=\textwidth]{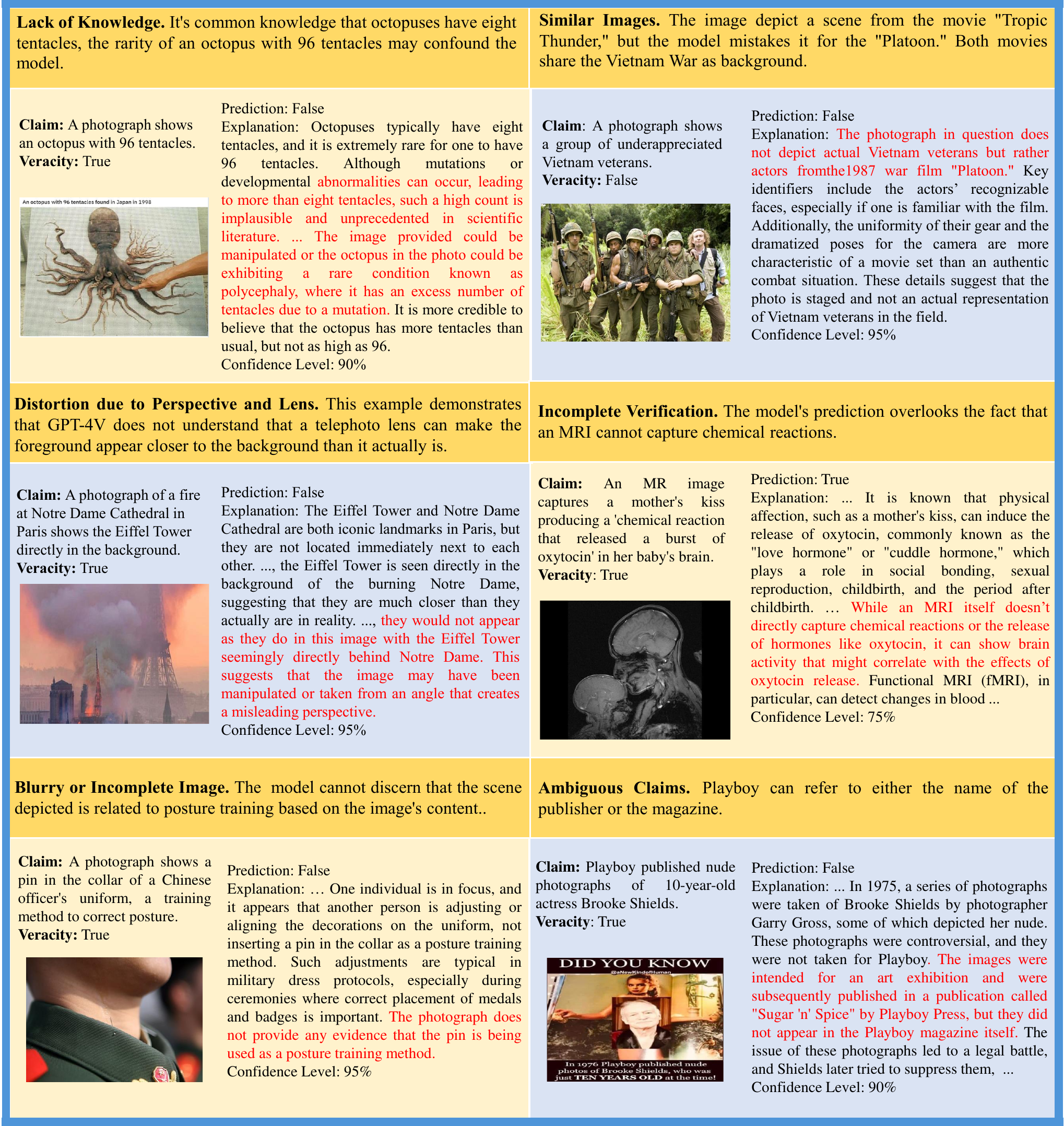}
    \caption{Taxonomy of failure reasons based on GPT-4V's responses. Each block includes the response from GPT-4V, along with explanation and reason for its failure.}
    \label{fig:taxnonomy}
\end{figure*}

\subsubsection{Analysis of Explanation Length}
For convenience, we refer to the number of sentences in the explanation part of a response as the explanation length. The average length of model explanations with different approaches is shown in Figure~\ref{fig:sent_nums}. We use NLTK to segment the sentences. GPT-4V provides the highest number of sentences in explanations across various datasets, with a minimum of 3.57 sentences in COSMOS and a maximum of 3.88 in Fauxtography. The value of LLaVA+PE represents the average length of explanations under different prompts. ICL prompts LLaVA to generate more sentences; however, providing two examples (LLaVA+ICL-2) does not yield longer explanations than providing just one (LLaVA+ICL-1), the average length even decreases.

We further analyze the relationship between the model's prediction accuracy and the explanation length, as depicted in Figure~\ref{fig:acc_vs_length}, where we display explanation lengths ranging from 2 to 7, encompassing the most samples generated by various models. Typically, different models achieve peak accuracy when the explanation length is 5 or 6 sentences. Specifically, on COSMOS, the accuracy of the different models grows steadily as the explanation length increases from 2 to 5. Across the other three datasets, the left half of the curves all show varying degrees of ascent despite noticeable fluctuations. GPT-4V achieves its highest accuracy at an explanation length of 5 sentences on them.

\input{results/04_table_noimage}

\subsubsection{Reasons for Predicting False}
To better understand the model's fact-checking process, particularly why it labels multimodal claims as false, we explore two types of cases: checked cases, the model confirms that statements have been verified by organizations; manipulated cases, it indicates that images have been altered.

\paragraph{Checked Cases} We count the occurrences of phrases such as "been checked" and "been debunked" across different settings, with the results displayed in left graph in Figure~\ref{fig:keywords}.  GPT-4V mentions these keywords most frequently across three public datasets, significantly more than others. Additionally, ICL prompts LLaVA to mention them more often. Notably, our ICL examples do not include checked cases. However, on Post-4V, several claims are considered fact-checked by LLaVA, but not by GPT-4V.  The accuracy of GPT-4V and LLaVA+ICL on checked cases is $86.5\%$ and $88.8\%$, respectively. The average True\&False Accuracy for those that are not checked cases is only 84.4\% and 65.7\%, respectively.


\paragraph{Manipulated Cases} We notice that only the examples in the 3\textsuperscript{rd} demonstration do not include manipulated cases. Therefore, we compare the impact of demonstrations on the number of manipulated cases, as shown in the right of Figure~\ref{fig:keywords}. Despite not accessing manipulated cases as example, the number significantly rises in demo3-1-exp1 and demo3-exp2, although these figures remain lower compared to other demonstrations. Thus, reasoning from the perspective of image manipulation is also not an explicit learning process. The accuracy of LLaVA is 67.6\% for manipulated cases and 65.4\% for non-manipulated cases, respectively. This demonstrates LLaVA's potential to analyze veracity from the perspective of image manipulation.


\input{results/04_tabe_reasoning_first}

\subsubsection{Taxonomy of Failure Reasons}
Upon examining GPT-4V's explanations, we identify six primary categories of failure reasons. \textit{Lack of Knowledge}, especially obscure knowledge that contradicts intuition, easily leads the model to assess a statement's truthfulness inaccurately.  \textit{Similar Images} refer to instances where the model associates with another image featuring a similar scene. In such cases, the model incorrectly extracts knowledge, leading to inconsistent content with the presented image.  \textit{Distortion Due to Perspective and Lens}, we find that GPT-4 cannot correctly understand images that are distorted due to the shooting angle or lens.. \textit{Incomplete Verification} refers to the cases where the model does not fully consider the atomic claims within a multimodal claim. For example, when a claim contains both correct and incorrect subclaims, or when the textual claim is accurate but the accompanying image is misleading. \textit{Blurry or Incomplete Image} can lead to an incorrect prediction due to the model's uncertainty about what it sees. \textit{Ambiguity Claims} involve semantic uncertainties, where a term could reference multiple entities. We show representative examples and explanations for each failure reason in Figure~\ref{fig:taxnonomy}. Our taxonomy can lead to targeted optimization strategies. For instance, we could include more images with distortions or blurry or incomplete images from image scaling and cropping into the instruction-tuning dataset.





\subsection{Ablation Studies}
\label{sec:ablation}
\paragraph{Impact of Images}



To understand the impact of images on fact-checking, we conduct a comparative experiment where we ask GPT-4V and LLaVA (13b) to re-evaluate the veracity of textual claims without providing images. As shown in Table~\ref{tab:noimage}, the results are marked in red for values that exceed those in Table~\ref{tab:comprehensive}. GPT-4V experiences a decrease in accuracy across all datasets, while surprisingly, LLaVA (13b) shows higher accuracy compared to when images are included. To understand this, we further look into their explanations.
As shown in Figure~\ref{fig:part-examples}, in the 1\textsuperscript{st} and 3\textsuperscript{rd} cases, the model can make correct predictions based on internal knowledge without images, stating "it is unlikely that Barack Obama would be endorsing his wife’s ..." and "Joe Biden is a public figure and it is highly unlikely that he would pose menacingly ..." However, the model is misled when manipulated images are presented. The GPT-4V's responses without image input are in Figure~\ref{fig:examples} in Appendix.



\paragraph{Reasoning First}
In the experiments above, we prompt the model to first output a prediction, followed by step-by-step reasoning. We conduct a comparative experiment to assess the impact of reversing this order—starting with reasoning before prediction—on accuracy.  The results are presented in Table~\ref{tab:reason_first}, where values exceeding those in Table~\ref{tab:comprehensive} are highlighted in red. We observe that starting with reasoning significantly increases the instances of GPT-4V expressing uncertainty. For example, the number of uncertain cases rises from 29 to 72 on Fauxtography and 10 to 38 on Post-4V. This increase directly leads to an obvious decline in Overall Accuracy, although True\&False Accuracy remains relatively high. 
LLaVA(7b) still predicts most claims as True. LLaVA(13b) shows varying accuracy improvements across the last three datasets, though the changes are not substantial. We also note an increase in instances where the model fails to follow instructions, which may partially influence the accuracy metrics.


%% file: results/04_table_noimage.tex
\begin{table*}[t]
\centering
\scriptsize
\setlength{\tabcolsep}{3pt}
\begin{tabular}{p{1.7cm}p{0.35cm}<{\centering}p{0.35cm}<{\centering}p{0.2cm}<{\centering}p{0.3cm}<{\centering}>{\columncolor{lightgray}}p{0.45cm}<{\centering}>{\columncolor{lightgray}}p{0.5cm}<{\centering}|p{0.35cm}<{\centering}p{0.35cm}<{\centering}p{0.2cm}<{\centering}p{0.3cm}<{\centering}>{\columncolor{lightgray}}p{0.45cm}<{\centering}>{\columncolor{lightgray}}p{0.5cm}<{\centering}|p{0.35cm}<{\centering}p{0.35cm}<{\centering}p{0.2cm}<{\centering}p{0.3cm}<{\centering}>{\columncolor{lightgray}}p{0.45cm}<{\centering}>{\columncolor{lightgray}}p{0.5cm}<{\centering}|p{0.35cm}<{\centering}p{0.35cm}<{\centering}p{0.2cm}<{\centering}p{0.3cm}<{\centering}>{\columncolor{lightgray}}p{0.45cm}<{\centering}>{\columncolor{lightgray}}p{0.5cm}<{\centering}}
\toprule
     & \multicolumn{6}{c}{\textbf{Fauxtography}} & \multicolumn{6}{c}{\textbf{COSMOS}} & \multicolumn{6}{c}{\textbf{MOCHEG}} & \multicolumn{6}{c}{\textbf{Post-4V}} \\ \midrule 
  & T    & F    & U    & O   & \cellcolor{white}{T\&F} & \cellcolor{white}{All}  & T   & F   & U  & O  & \cellcolor{white}{T\&F} & \cellcolor{white}{All}  & T   & F   & U  & O  & \cellcolor{white}{T\&F} & \cellcolor{white}{All}   & T   & F  & U  & O  & \cellcolor{white}{T\&F} & \cellcolor{white}{All} \\
\midrule
 GPT-4V     &  105    & \textcolor{red}{202}   &   \textcolor{red}{32}  &  \textcolor{red}{61} &  80.1 & 71.0   & 111  &  \textcolor{red}{253}   & 50  & \textcolor{red}{66} &  64.0 & 57.5  &  153   & \textcolor{red}{265}    & \textcolor{red}{38}  & \textcolor{red}{48}   &  73.9 & 68.9  & 35     &  68   &  10   &  \textcolor{red}{73}  &      74.8 & 67.7  \\
 LLaVA(7b)     &   \textcolor{red}{398}   &  0    &   \textcolor{red}{2} &   0  &   50.3 & \textcolor{red}{50.0}    & \textcolor{red}{474}    &  0   &  4  &  2   &  50.2 & 49.8    &  \textcolor{red}{502}   &   0  &  \textcolor{red}{2}  &  0  &   53.0 & \textcolor{red}{52.8}   &  \textcolor{red}{181}   &  0  &  \textcolor{red}{4}  &  1  &  \textcolor{red}{43.1} & \textcolor{red}{43.0}    \\
 LLaVA(13b)     &  167    & \textcolor{red}{214}     & \textcolor{red}{17}    &  2   &  \textcolor{red}{62.5} & \textcolor{red}{62.0}     &   263  &  \textcolor{red}{195}   & \textcolor{red}{20}   &  2  &  \textcolor{red}{76.6} & \textcolor{red}{75.8}    &  168   &  \textcolor{red}{316}   &  20  &  0  &   \textcolor{red}{62.0} & \textcolor{red}{61.5}   &  40   & \textcolor{red}{124}   &  \textcolor{red}{22}  &  0  &   \textcolor{red}{60.4} & \textcolor{red}{60.8}   \\
\bottomrule
\end{tabular}
\caption{Performance of various models without image input. Red numbers highlight those have increased compared to those with images. Text-only claims decrease the accuracy of GPT-4V but increase the accuracy of LLaVA (13b).}
\label{tab:noimage}
\end{table*}

%% file: results/04_tabe_reasoning_first.tex
\begin{table*}[t]
\centering
\scriptsize
\setlength{\tabcolsep}{3pt}
\begin{tabular}{p{1.7cm}p{0.35cm}<{\centering}p{0.35cm}<{\centering}p{0.2cm}<{\centering}p{0.3cm}<{\centering}>{\columncolor{lightgray}}p{0.45cm}<{\centering}>{\columncolor{lightgray}}p{0.5cm}<{\centering}|p{0.35cm}<{\centering}p{0.35cm}<{\centering}p{0.2cm}<{\centering}p{0.3cm}<{\centering}>{\columncolor{lightgray}}p{0.45cm}<{\centering}>{\columncolor{lightgray}}p{0.5cm}<{\centering}|p{0.35cm}<{\centering}p{0.35cm}<{\centering}p{0.2cm}<{\centering}p{0.3cm}<{\centering}>{\columncolor{lightgray}}p{0.45cm}<{\centering}>{\columncolor{lightgray}}p{0.5cm}<{\centering}|p{0.35cm}<{\centering}p{0.35cm}<{\centering}p{0.2cm}<{\centering}p{0.3cm}<{\centering}>{\columncolor{lightgray}}p{0.45cm}<{\centering}>{\columncolor{lightgray}}p{0.5cm}<{\centering}}
\toprule
     & \multicolumn{6}{c}{\textbf{Fauxtography}} & \multicolumn{6}{c}{\textbf{COSMOS}} & \multicolumn{6}{c}{\textbf{MOCHEG}} & \multicolumn{6}{c}{\textbf{Post-4V}} \\ \midrule 
  & T    & F    & U    & O   & \cellcolor{white}{T\&F} & \cellcolor{white}{All}  & T   & F   & U  & O  & \cellcolor{white}{T\&F} & \cellcolor{white}{All}  & T   & F   & U  & O  & \cellcolor{white}{T\&F} & \cellcolor{white}{All}   & T   & F  & U  & O  & \cellcolor{white}{T\&F} & \cellcolor{white}{All} \\
\midrule
 GPT-4V     &  158 & 158  &   \textcolor{red}{72}  &  12 &  80.7 & 63.8   & 162  &  190   & \textcolor{red}{109} & \textcolor{red}{18} &  79.8 & 70.0  &  \textcolor{red}{234}   & 204    & \textcolor{red}{54}  & \textcolor{red}{33}  &  83.7 & 77.8  & 51  &  83  &  \textcolor{red}{38}  &  \textcolor{red}{14}  &  \textcolor{red}{80.1} & 72.6 \\
 LLaVA(7b)     &  237  &  \textcolor{red}{71} & \textcolor{red}{7} &   49  &   \textcolor{red}{54.1} & \textcolor{red}{50.5}    & 311    & \textcolor{red}{101}   &  \textcolor{red}{14}  &  \textcolor{red}{54}   &  \textcolor{red}{66.0} & \textcolor{red}{60.1}    &  341   &  \textcolor{red}{89}  &  \textcolor{red}{8}  &  66 &   \textcolor{red}{56.0} & \textcolor{red}{50.8}   &  125   &  \textcolor{red}{37}  &  3  &  21  &  \textcolor{red}{51.2} & \textcolor{red}{49.5}    \\
 LLaVA(13b)  &  207    & \textcolor{red}{143}     & \textcolor{red}{7}    &  \textcolor{red}{42}   &  51.7 & 52.0     &   228  &  \textcolor{red}{201}   &  \textcolor{red}{22}  &  \textcolor{red}{28}  &  \textcolor{red}{69.9} & \textcolor{red}{68.5}    &  235   &  \textcolor{red}{212}   &  7  &  \textcolor{red}{50}  &   \textcolor{red}{61.8} & \textcolor{red}{59.5}   &  65   & \textcolor{red}{105}   &  2  &  \textcolor{red}{14}  &  \textcolor{red}{61.8} & \textcolor{red}{60.2} \\  
\bottomrule
\end{tabular}
\caption{Performance of various models when reasoning first. Red numbers highlight those that have increased compared to those predicting first. Reasoning first significantly increases the expression of uncertainty in GPT-4V.}
\label{tab:reason_first}
\end{table*}

%% file: sections/06_limitations.tex
\section{Limitations}
We acknowledge the following limitations in our work. First, due to constraints in manpower and computational resources, we did not test the complete Fauxtography and COSMOS datasets. The limited number of samples may impede an accurate assessment of model characteristics. Second, we did not quantify the different capabilities within fact-checking, such as the ability of different models to recognize AI-generated images, identify faces and scenes, and retrieve contextual knowledge. Moreover, we did not investigate how fine-tuning with domain-specific data could potentially improve model performance. Future research will aim to address these gaps and provide a more comprehensive evaluation of the models.




%% file: sections/05_conclustion.tex
\section{Conclusion and Future Work}\label{sec:conclusion}
We investigated the capabilities of MLLMs to fact-check real-world claims, relying solely on their parametric knowledge and reasoning capabilities without external references. We proposed an evaluation framework, designed various experiments to address the research questions. Our results indicated that state-of-the-art MLLMs, such as GPT-4V, have the potential to assist professional fact-checkers: they can enhance their efficiency by providing reference predictions, valuable clues, and explanatory insights, together with confidence. However, open-source models like LLaVA fall behind. They can be misled by manipulated images and may generate hallucinations when no image is present. We empirically demonstrate that prompt ensembles (PE) and in-context learning (ICL) are able to improve model's accuracy in detecting misinformation, while ICL is more efficient. 


In future work, we plan to investigate how to enhance the fact-checking capabilities of models, focusing not only on improving accuracy but also on strengthening robustness and augmenting the knowledge reasoning and the verification abilities required for fact-checking. A potential approach is to use GPT-4V to enhance the capabilities of smaller open-source models through knowledge distillation. We also plan to explore how MLLMs can better support fact-checking when connected to external knowledge.

%% file: sections/07_appendix.tex
\input{results/07_pe_table}
\input{results/07_icl_table}

\section{More Experimental Results}
\begin{figure*}[!htbp]
    \centering
    \includegraphics[width=0.95\textwidth]{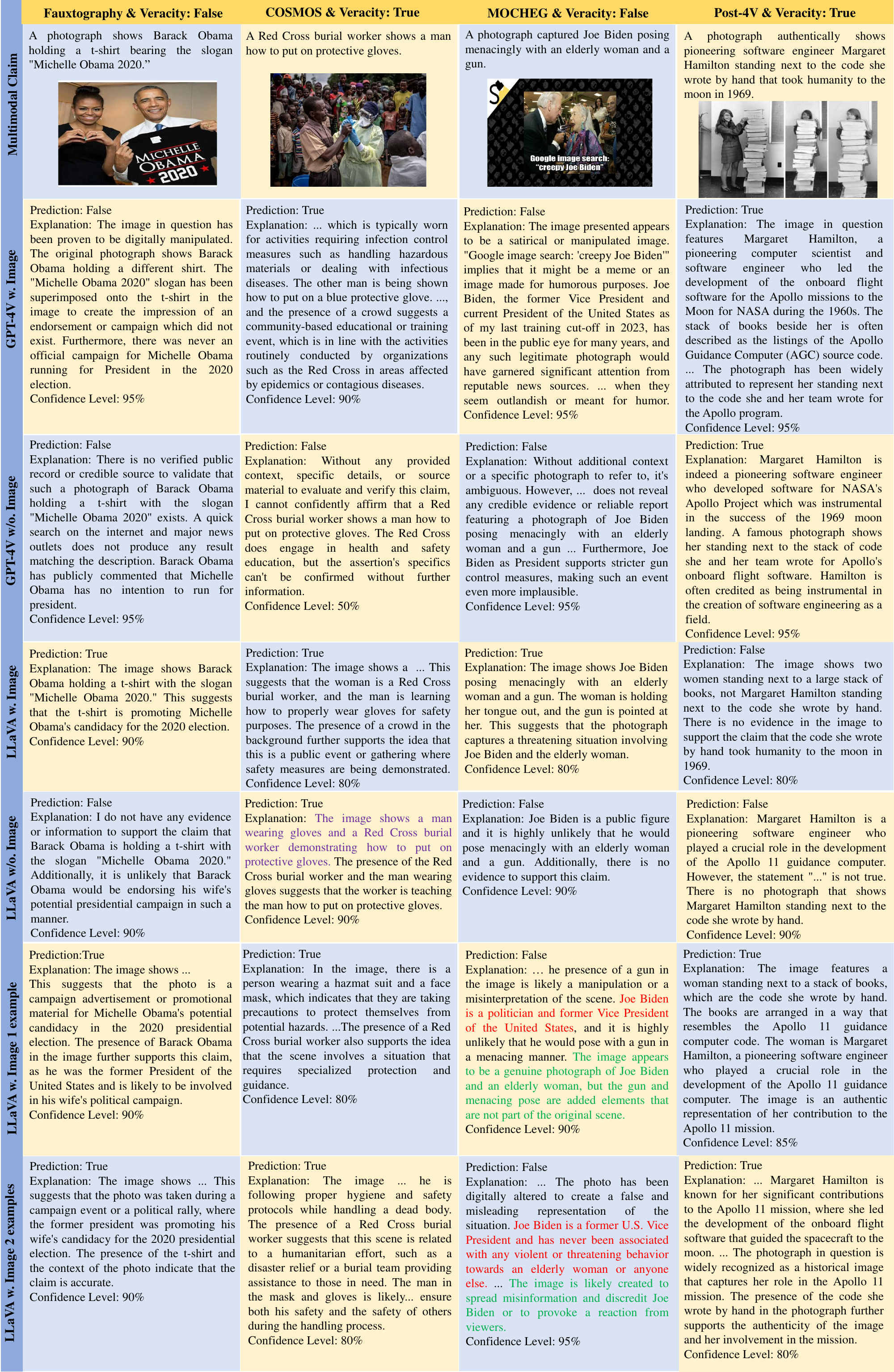}
    \caption{More detailed responses from GPT-4V and LLaVA(13b) across various experimental setups. Purple text indicates hallucinations by the model when no images are present, red text shows outdated knowledge, and green text displays the model's analysis of image manipulation.}
    \label{fig:examples}
\end{figure*}

Figure~\ref{fig:examples} displays additional responses, where "w. Image" indicates multimodal claims as input and "w/o. Image" denotes text-only claims as input. The last two rows show the results for LLaVA+ICL-1 and LLaVA+ICL-2, respectively. As shown in Figure~\ref{fig:examples}, GPT-4V's explanations demonstrate its awareness of missing images, employing phrases such as "there is no verified public record" or "without any provided context." This indicates that GPT-4V actively retrieves knowledge to verify the presence or context of images. Compared with no demonstrations, the results reveal that ICL enhances the model's reasoning capabilities, incorporating more detailed analysis of images and discussions about the motivations behind their creation.

Table~\ref{tab:pe} shows the results with different prompts in prompt ensembles (PE). The accuracy of the GPT-4V model is quite consistent across various prompts. In contrast, the LLaVA model is very sensitive to the prompts, with some prompts showing significantly higher accuracy than others on the same dataset. However, no single prompt consistently performs best across all datasets. Table~\ref{tab:icl} displays the impact of different demonstrations in ICL on the LLaVA (13b) model. Demonstrations clearly enhance the model's accuracy. However, the benefit of the second example is limited, and it does not show a significant improvement compared to the ICL-I cases.

\begin{figure*}[!htbp]
    \centering
    \includegraphics[width=\textwidth]{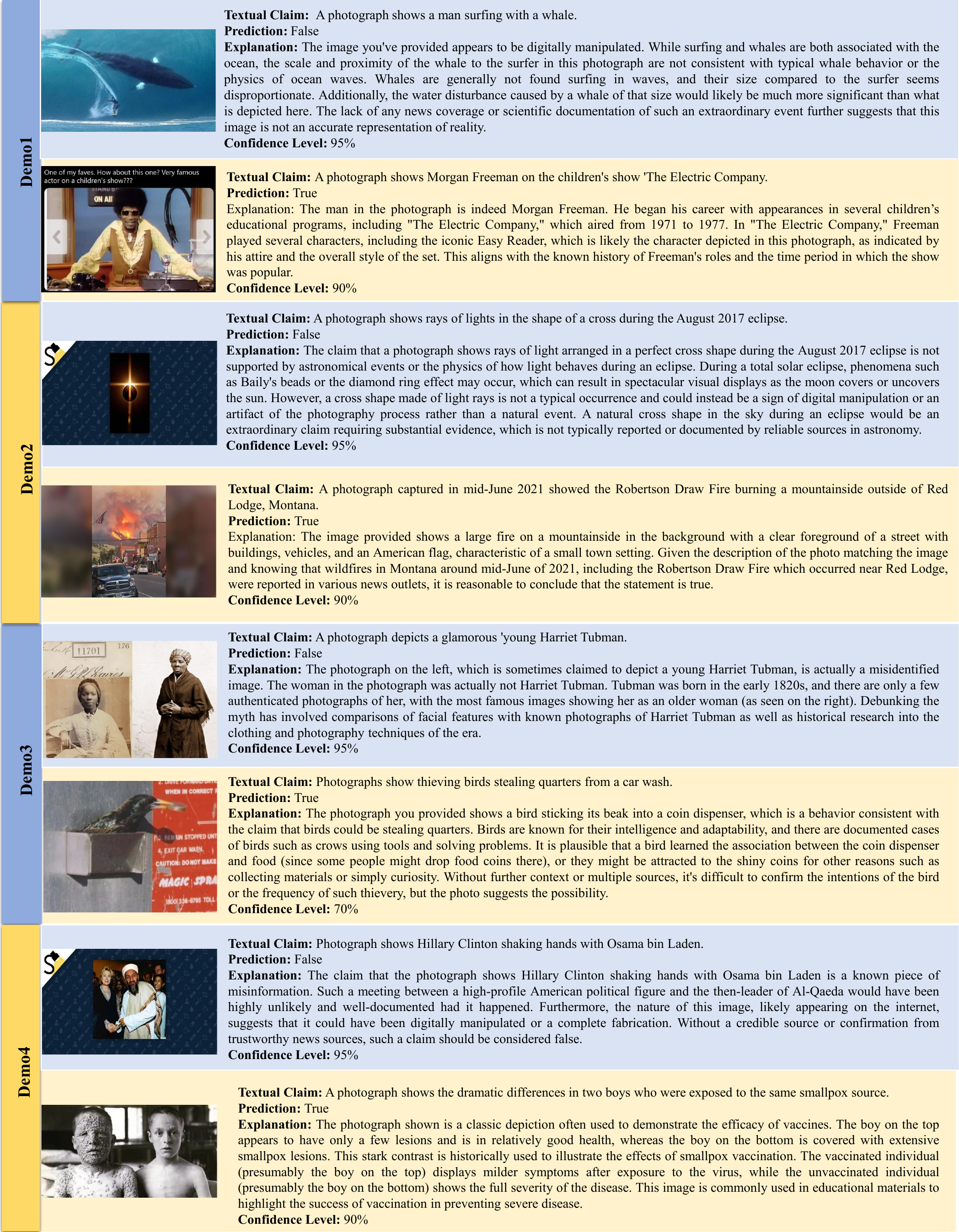}
    \caption{Four demonstration pairs used in the ICL experiments, ICL-1 uses only the first example from each demonstration, while ICL-2 uses both examples. They are collected from GPT-4V and factually correct.}
    \label{fig:icl}
\end{figure*}



%% file: results/07_pe_table.tex
\begin{table*}[t]
\centering
\scriptsize
\setlength{\tabcolsep}{5pt}
\begin{tabular}{p{0.4cm}<{\centering}p{0.5cm}<{\centering}p{0.2cm}<{\centering}p{0.2cm}<{\centering}p{0.1cm}<{\centering}p{0.2cm}<{\centering}>{\columncolor{lightgray}}p{0.4cm}<{\centering}>{\columncolor{lightgray}}p{0.4cm}<{\centering}|p{0.2cm}<{\centering}p{0.2cm}<{\centering}p{0.1cm}<{\centering}p{0.2cm}<{\centering}>{\columncolor{lightgray}}p{0.4cm}<{\centering}>{\columncolor{lightgray}}p{0.4cm}<{\centering}|p{0.2cm}<{\centering}p{0.2cm}<{\centering}p{0.1cm}<{\centering}p{0.2cm}<{\centering}>{\columncolor{lightgray}}p{0.4cm}<{\centering}>{\columncolor{lightgray}}p{0.4cm}<{\centering}|p{0.2cm}<{\centering}p{0.1cm}<{\centering}p{0.1cm}<{\centering}p{0.2cm}<{\centering}>{\columncolor{lightgray}}p{0.4cm}<{\centering}>{\columncolor{lightgray}}p{0.4cm}<{\centering}}
\toprule
  &    & \multicolumn{6}{c}{\textbf{Fauxtography}} & \multicolumn{6}{c}{\textbf{COSMOS}} & \multicolumn{6}{c}{\textbf{MOCHEG}} & \multicolumn{6}{c}{\textbf{Post-4V}} \\ \midrule 
& idx & T    & F    & U    & O   & T\&F & All   & T   & F   & U  & O  & T\&F & All  & T   & F   & U  & O  & T\&F & All  & T   & F  & U  & O  & T\&F & All \\
\midrule
\multirow{3}{*}{GPT-4V} & 0     &  158    &   195   &   29  & 18 &  81.9 & \textbf{82.1}   & 179  &  204   & 83  & 14 &  86.2 & 80.0  &  216   & 223   & 37  & 28   &  \textbf{87.2} & \textbf{83.5}  & 54     &  98   &  26  &  8  &      \textbf{79.6} & 73.7  \\
&  2    & 158   &  181   &   32  & 29  & \textbf{84.1}  & 81.3  & 163   & 219   & 75  & 23 & 85.1 & 80.8  & 205  & 234  & 34 & 31 & \textbf{82.1} & 81.2 &  55&     95 &     22&  14 &     77.8 & \textbf{75.3}   \\
& 5 &  143    & 202  &  31   &  24 & 81.2  & 76.3  & 175  & 216    &  72 & 17 &  \textbf{88.0} & \textbf{81.5}  & 179 & 250 & 44 & 31    & 78.6  & 75.6 &   51   &  106   &  18   &  11 &       76.4 & 71.5   \\
\midrule
\multirow{6}{*}{LLaVA} & 0      &  286    &   98   &   3  & 13 &  54.4 & 52.3   & 331  &  134   & 13  & 2 &  69.7 & 67.7  &  363   & 112    & 28 & 1    &  61.5 & 58.1 & 116     &  65   &  2   &  3  &      56.3 & 55.9   \\
& 1      &     197 &   163   &  7    &  33   &   48.6 & 48.5     &  169  &  236   &  15  &  60  &    69.6 & 66.3   &  247   &  211  &  9  &     37 &  57.6 & 57.9   &  68  &  105  &  5  &  8 & 56.6 & \textbf{57.5}    \\
& 2  &  170   &   223   &  4    &   3   &   49.1 & 49.5  &   197    &  274   &  7   &  2  &  69.0 & 68.5  &  318    &  178   &   7  &  1  &  58.1 & 57.9  &   41   &  141   &   4 &   0 &   56.0 & 57.0      \\
& 3      &  243     &  145   &   8   &  4   &   52.6 & 53.0   &  243   &  222    &  13  &  2  &   71.8 & 70.2   &   295  &   197  &   11 &  1  &    57.3 & 56.3  &   82  &   93 &  10  &  1  &   56.5  & 56.5   \\
& 4      &  211    &  180    &   6   &  3   &  62.1 & 62.0      &   232  &  244    &  3  &  1  &   77.3 & 77.1   &  210   &  286   &   4 &  4  &   57.3 &  56.3  &   55  &  129  &  2  &  0  &   \textbf{57.6} & 57.0   \\
& 5    &   241   &   154   &  3    &  2   &   \textbf{63.3} & \textbf{63.0}    &  313   &  163  &  4  &   0   &  \textbf{80.5} & \textbf{80.4}  &  260   &  236  &   8 &  0    &  \textbf{62.1} & \textbf{62.3 }  & 109   &  75  & 2   &  0  & 52.7 & 52.7  \\
\bottomrule
\end{tabular}
\caption{Performance of GPT-4V and LLaVA(13b), with the use of different prompts. Prompt 0 represents the original prompt in Section~\ref{sec:eval_prompt}; others are suggested by ChatGPT, as shown in Figure~\ref{fig:prompts}.}
\label{tab:pe}
\end{table*}

%% file: results/07_icl_table.tex
\begin{table*}[t]
\centering
\setlength{\tabcolsep}{5pt}
\scriptsize
\begin{tabular}{p{0.5cm}<{\centering}p{0.4cm}<{\centering}p{0.2cm}<{\centering}p{0.2cm}<{\centering}p{0.1cm}<{\centering}p{0.2cm}<{\centering}>{\columncolor{lightgray}}p{0.4cm}<{\centering}>{\columncolor{lightgray}}p{0.4cm}<{\centering}|p{0.2cm}<{\centering}p{0.2cm}<{\centering}p{0.1cm}<{\centering}p{0.2cm}<{\centering}>{\columncolor{lightgray}}p{0.4cm}<{\centering}>{\columncolor{lightgray}}p{0.4cm}<{\centering}|p{0.2cm}<{\centering}p{0.2cm}<{\centering}p{0.1cm}<{\centering}p{0.2cm}<{\centering}>{\columncolor{lightgray}}p{0.4cm}<{\centering}>{\columncolor{lightgray}}p{0.4cm}<{\centering}|p{0.2cm}<{\centering}p{0.1cm}<{\centering}p{0.1cm}<{\centering}p{0.2cm}<{\centering}>{\columncolor{lightgray}}p{0.4cm}<{\centering}>{\columncolor{lightgray}}p{0.4cm}<{\centering}}
\toprule
&      & \multicolumn{6}{c}{\textbf{Fauxtography}} & \multicolumn{6}{c}{\textbf{COSMOS}} & \multicolumn{6}{c}{\textbf{MOCHEG}} & \multicolumn{6}{c}{\textbf{Post-4V}} \\ \midrule 
Demo & Num & T    & F    & U    & O   & T\&F & All   & T   & F   & U  & O  & T\&F & All  & T   & F   & U  & O  & T\&F & All  & T   & F  & U  & O  & T\&F & All \\
\midrule
0 & 0      &  286    &   98   &   16  & 0 &  54.4 & 52.8   & 331  &  134   & 14  & 1 &  69.7 & 67.7  &  363   & 112    & 29& 0    &  61.5 & 58.1 & 116     &  65   &  5   &  0  &      56.4 & 55.9   \\ 
 \midrule
\multirow{2}{*}{1} & 1      &  244    &   135   &   9  & 12 &  60.9 & 61.8   & 295  &  171   & 7  & 7 &  75.3 & 74.2  &  255   & 229    & 12 & 8   &  62.6 & 62.7  & 95     &  88   &  3   &  0  &      56.3 & 55.4  \\
&  2     &   182   &  161    &   6  &   51  &   59.5 & 60.0    & 234    &  220   &  8  &  18   &  75.8 & 73.5    &  169   &   282  &  10  &  43  &   57.4 & 55.5   &  42   &  109  &  6  &  29  &  \textbf{67.5} & \textbf{63.4}    \\
 \midrule
\multirow{2}{*}{2} & 1     &  264    & 121     & 4     &  11   &  61.6 & 62.8     &   337  &  131   & 7   &  5  &  73.9 & 73.5    &  314   &  180   &  4  &  6  &   \textbf{64.8} & \textbf{65.1}   &  97   & 79   &  10  &  0  &   57.4 & 55.9   \\
&  2      &  176    & 204     & 10     &  10   &   60.0 & 60.5   &   233  &   230  &   11 &  6  &   \textbf{80.3} & \textbf{79.6}   &   199  &  291   & 8   & 6   &  61.2 & 60.7  &  44   &   131 &  11  &  0  &  60.0 & 60.8   \\
\midrule
\multirow{2}{*}{3} & 1      &  211    &  185    &  1    &   3  &   62.6 & 62.8   &  279   & 195   &   6 &   0   &  72.2 & 71.7   &  236   &  266   &  2  &  0    &  61.8 & 61.5  &   75 &  111  &  0  &   0  & 60.2 & 60.2 \\
&  2      &   191   &  198    &  9    &  2   &  \textbf{64.1} & \textbf{63.8}     &   263  &   201  &  16  & 0   &   75.4 & 75.4    & 206    &  288  & 9   &  1    &  61.1 & 61.7   &  54   & 127   &  4   &  1 & 59.7 & 59.7  \\
 \midrule
\multirow{2}{*}{4} & 1      &  191   &  193    &  9    &   7  &   62.2 & 62.0   &  260   & 203   &   16 &   1   &  78.0 & 77.9  &  210  &  283   &  11  &  0    &  60.6 & 60.9  &   50 &  126  &  8  &   2  & 60.8 & 60.8 \\
&  2      &   196  &  188   &  7   &  9   &  62.8 & 62.8     &   259  &   207  &  12  & 2   &   77.5 & 77.5    & 207    &  284  & 11   &  2    &  61.1 & 61.3   &  52   & 122   &  10   &  2 & 61.5 & 61.8 \\
\bottomrule
\end{tabular}
\caption{Detailed performance of LLaVA(13b) when demonstrations are presented. \textit{Demo: demonstration index, Num: number of examples in the demonstration.} The first row, with no demonstration, establishes the baseline. The table presents outcomes from four distinct demonstration pairs, each comprising a true claim followed by a false one.}
\label{tab:icl}
\end{table*}